\documentclass[a4paper, 10pt, conference]{ieeeconf}      % Use this line for a4 paper

\IEEEoverridecommandlockouts                              % This command is only needed if 
\usepackage{amsmath}
\usepackage{amssymb}
\usepackage{makecell}
\usepackage{ragged2e}
\usepackage{mathrsfs}
\usepackage{epsfig}
\usepackage{graphicx}
\usepackage{geometry}
\graphicspath{{Images/}}
\DeclareGraphicsExtensions{.pdf, .png, .jpg, .eps}

\usepackage{multirow} 
\usepackage{multicol} 
\usepackage[colorinlistoftodos]{todonotes}
\usepackage{array, textcomp, stackrel, url, mathtools, enumerate}

\usepackage{float}
\usepackage{subfig}
\usepackage{footnote}
\usepackage{tablefootnote}
\usepackage{color}
\usepackage{graphicx}
\usepackage{colortbl}
\usepackage{comment}
\usepackage{cite}
\usepackage{amsmath}
\usepackage{caption}
\usepackage{arydshln}
\usepackage[pagewise]{lineno}

\title{\LARGE \bf
Lightweight Monocular Depth Estimation with an Edge Guided Network
}

\author{Xingshuai Dong$^{1}$, Matthew A. Garratt$^{1}$, Sreenatha G. Anavatti$^{1}$, Hussein A. Abbass$^{1}$ and Junyu Dong$^{2}$% <-this % stops a space
\thanks{$^{1}$Xingshuai Dong, Matthew A. Garratt, Sreenatha G. Anavatti and Hussein A. Abbass are with the School of Engineering and Information Technology, University of New South Wales, Canberra, ACT 2612, Australia }%
\thanks{$^{2}$Junyu Dong is with the School of Computer Science and Technology, Ocean University of China, Qingdao 266100, China}%
}

\begin{document}

\maketitle
\thispagestyle{empty}
\pagestyle{empty}

%%%%%%%%%%%%%%%%%%%%%%%%%%%%%%%%%%%%%%%%%%%%%%%%%%%%%%%%%%%%%%%%%%%%%%%%%%%%%%%%
\begin{abstract}
Monocular depth estimation is an important task that can be applied to many robotic applications. Existing methods focus on improving depth estimation accuracy via training increasingly deeper and wider networks, however these suffer from large computational complexity. Recent studies found that edge information are important cues for convolutional neural networks (CNNs) to estimate depth. Inspired by the above observations, we present a novel lightweight \underline{E}dge \underline{G}uided \underline{D}epth Estimation \underline{Net}work (EGD-Net) in this study. In particular, we start out with a lightweight encoder-decoder architecture and embed an edge guidance branch which takes as input image gradients and multi-scale feature maps from the backbone to learn the edge attention features. In order to aggregate the context information and edge attention features, we design a transformer-based feature aggregation module (TRFA). TRFA captures the long-range dependencies between the context information and edge attention features through cross-attention mechanism. We perform extensive experiments on the NYU depth v2 dataset. Experimental results show that the proposed method runs about 96 fps on a Nvidia GTX 1080 GPU whilst achieving the state-of-the-art performance in terms of accuracy.
\end{abstract}

%%%%%%%%%%%%%%%%%%%%%%%%%%%%%%%%%%%%%%%%%%%%%%%%%%%%%%%%%%%%%%%%%%%%%%%%%%%%%%%%
\section{INTRODUCTION}
Depth estimation refers to estimating depth maps from RGB images, and has been widely explored in computer vision and robotics \cite{dong2022towards}. It is a fundamental perception component of robotic systems. Active sensors such as RGB-D cameras and LiDAR provide accurate depth perception. However, these devices are heavy and require high power consumption that cannot be deployed on resource constrained platforms. In contrast, monocular depth estimation (MDE) is inexpensive, and achievable in compact form factors with high energy efficiency. \par
Recent developed methods \cite{laina2016deeper, alhashim2018high, chen2019structure, hu2019revisiting, ye2021dpnet} learn deeper (more layers) and wider (more channels in each layer) models that produce better depth estimation performance. Particularly, these methods aim to improve depth estimation accuracy other than achieve real-time running speed. This makes them difficult to run in edge platforms where reaction time is crucial for operating safety such as for obstacle avoidance in small sized autonomous robots. \par
In order to solve the problem of running speed on embedded platforms, lightweight CNNs such as ERFNet \cite{romera2017erfnet} and MobileNets \cite{howard2017mobilenets, sandler2018mobilenetv2} have been employed to design lightweight MDE networks \cite{spek2018cream, wofk2019fastdepth}. Moreover, Wofk et al. \cite{wofk2019fastdepth} applied network pruning technique to further reduce the number of parameters. Although these methods achieve real-time speed on embedded platforms, their accuracy are inferior to state-of-the-art methods. \par
In addition, the above discussed methods \cite{spek2018cream, wofk2019fastdepth, rudolph2022lightweight} employ encoder-decoder style network architectures. However the downsampling operations in the encoder network distort fine details in lower resolution layers, which leads to blurry results around object edges. To avoid the loss of spatial information, MobileXNet \cite{dong2021mobilexnet} stacked two relatively shallow encoder-decoder style subnetworks back-to-back in a unified framework. According to \cite{dijk2019neural, hu2019visualization}, the edges in input images are important cue for CNNs to predict depth. \par
\begin{figure}
	\centering
    \includegraphics[width=.45\textwidth]{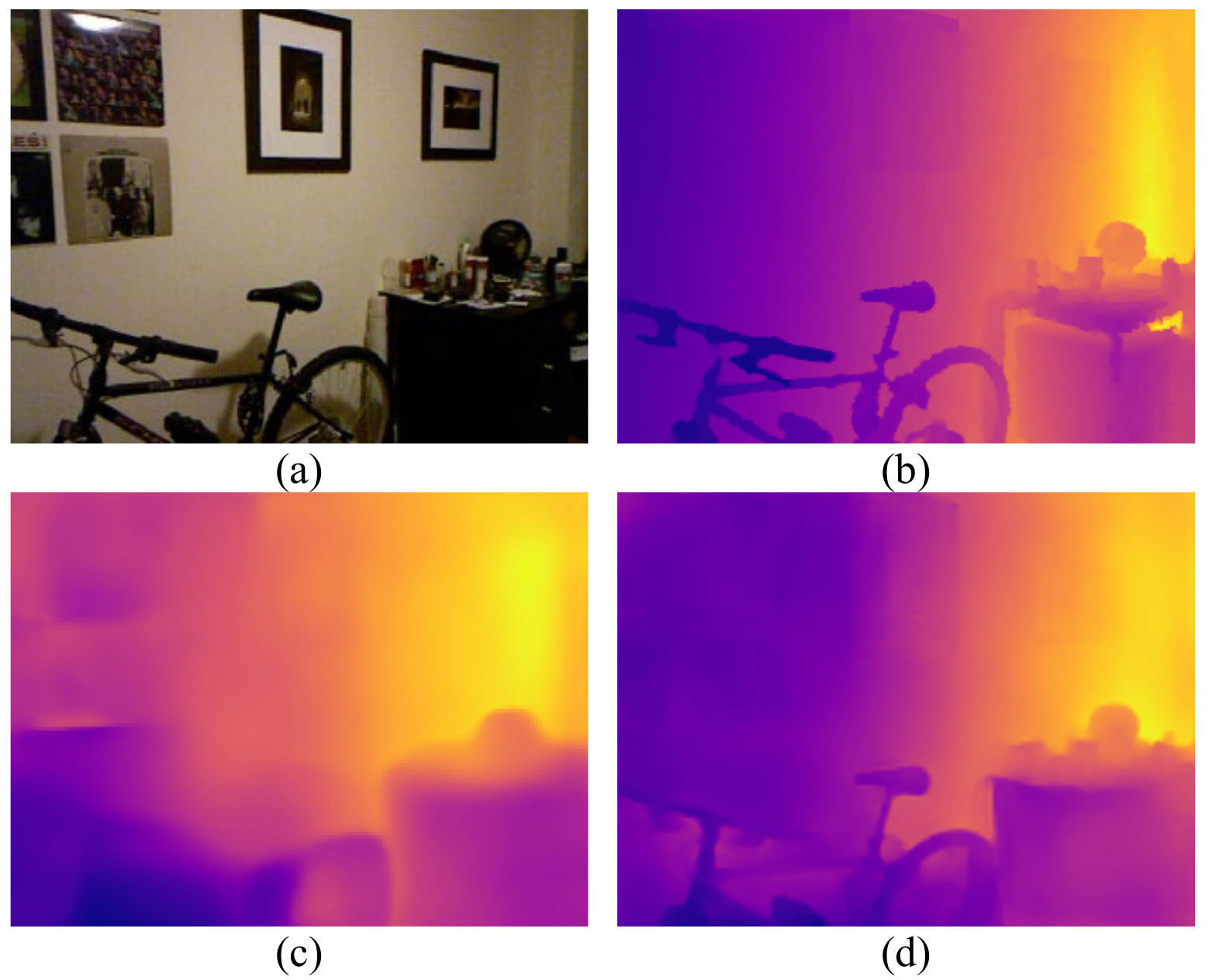}
    \caption{Example comparison of estimated depth maps. (a) RGB image, (b) Ground-truth depth, (c) Wofk et al. \cite{wofk2019fastdepth} and (d) Our method.}
    \label{fig:example_comparison}
\end{figure}
This study aims to address the above problems by designing a novel lightweight network which adopts edge attention features to guide MDE. Specifically, we integrate the depth estimation branch and an edge guidance branch in a unified network, named \underline{E}dge \underline{G}uided \underline{D}epth Estimation \underline{Net}work (EGD-Net). EGD-Net is built on top of a shallow (less layers) and narrow (less channels in each layer) encoder-decoder network and applies edge attention features to guide the depth estimation. The contributions of this study are attributed as follows: (1) we propose a novel lightweight monocular depth estimation network, named EGD-Net, which employs edge attention features to guide the task of depth estimation; (2) we design a channel attention-based feature fusion module; (3) we design a transformer-based feature aggregation module, which captures the long-range dependencies between the edge attention features and the context information; (4) we demonstrate the effectiveness of our designed method through extensive experiments. \par
\begin{figure*}
	\centering
    \includegraphics[width=.9\textwidth]{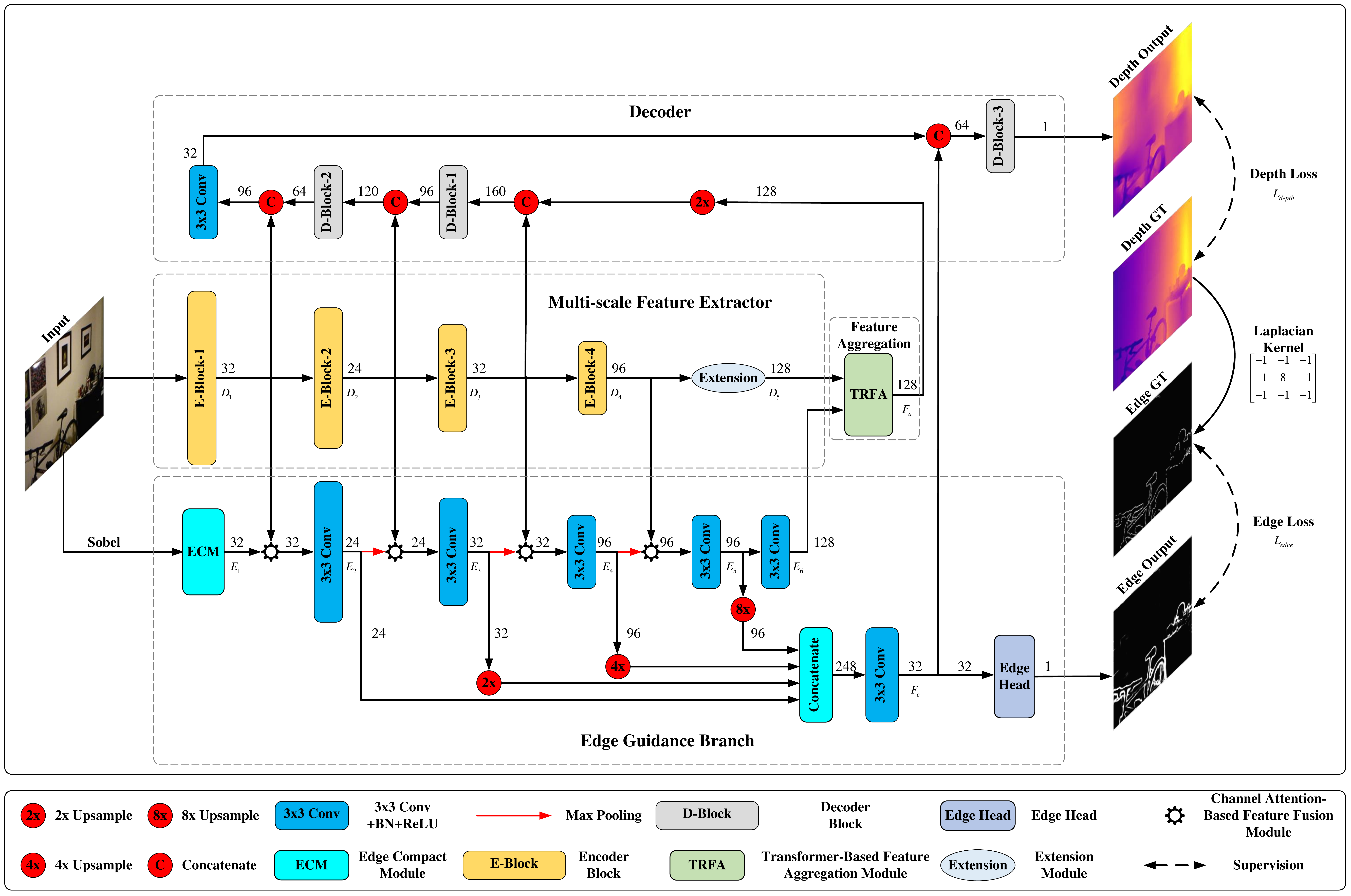}
    \caption{Illustration of our EGD-Net architecture.} 
    %Our architecture includes four parts, multi-scale feature extractor, edge guidance branch, feature aggregation module and decoder. The multi-scale feature extractor can be any backbone architecture. The edge guidance branch focuses on edge attention feature learning through a set of convolutional layers, feature fusion modules and supervision. The feature aggregation module aggregates feature representations from the multi-scale feature extractor and edge guidance branch.} 
    \label{fig:pipeline}
\end{figure*}
\section{Related Work}
\subsection{Convolutional Neural Networks}
Starting from the appearance of AlexNet \cite{krizhevsky2012imagenet}, CNNs have significantly advanced the progress of computer vision. Over the past few years, many CNN architectures such as ResNet \cite{he2016deep} and SENet \cite{hu2018squeeze} have been designed. The development trend is to increase the depth and width of CNNs to enhance the accuracy. However, these advances neglect the need to make networks more efficient with respect to the size and speed required for some real-time applications. Howard et al. \cite{howard2017mobilenets} designed the first lightweight CNN, MobileNet, for mobile and embedded vision applications. MobileNet is built on top of depthwise separable convolutions, which decompose a regular convolution into a depthwise convolution and a $1\times1$ pointwise convolution. Later, Sandler et al. \cite{sandler2018mobilenetv2} extended \cite{howard2017mobilenets} through a novel efficient convolutional block with inverted residual and linear bottleneck. ShuffleNet \cite{ma2018shufflenet} applied channel shuffle operators in the channel dimension of feature maps to make cross-group information flow for group convolution layers. Tan et al. \cite{tan2019efficientnet} developed an EfficientNet-B0 baseline network by using neural architecture search and scaled it up to get a set of models, named EfficientNets. \par
\subsection{Monocular Depth Estimation}
Monocular depth estimation (MDE) is a task that estimates dense depth maps from single RGB images. Motivated by the success of CNNs in image classification, Eigen et al. \cite{eigen2014depth} designed the first MDE network. Later, Laina et al. \cite{laina2016deeper} designed a fully convolutional network which includes an encoder and a decoder. Inspired by \cite{laina2016deeper}, Hu et al. \cite{hu2019revisiting} combined the encoder-decoder network with a multi-scale feature fusion module and a refinement module. Later, Chen et al. \cite{chen2019structure} exploited multiple scale scene structure information to estimate depth maps. Ye et al. \cite{ye2021dpnet} designed a dual branch network architecture for MDE, named DPNet. DPNet incorporates a spatial branch to retain spatial details and produce high resolution features. The produced high resolution features are fused with features from the contextual branch in a refinement module. Recently, Chang et al. \cite{chang2021transformer} designed a hybrid network consisting of a Transformer-based encoder and a CNN-based decoder to solve the MDE problem. The above discussed methods focus more on depth estimation accuracy and use large backbones or complex architectures to achieve performance increase which lead to increased computational complexity. \par
Spek et al. \cite{spek2018cream} designed a lightweight depth estimation network on the basis of ``non-bottleneck-1D" block \cite{romera2017erfnet}. Although the designed network achieves real-time speed on the Nvidia TX2 GPU, its accuracy is inferior. Wofk et al. \cite{wofk2019fastdepth} introduced a lightweight encoder-decoder network for MDE. Besides, they employed network pruning method to further reduce the amount of parameters. Dong et al. \cite{dong2021mobilexnet} introduced a real-time MDE network which stacks two simple encoder-decoder style network in a unified framework. Existing methods employ encoder-decoder network to aggregate high-level and low-level features to regress depth. Researches in \cite{dijk2019neural, hu2019visualization} demonstrated that the edges in input images are important cue for CNNs to predict depth. In addition, Fan et al. \cite{fan2021rethinking} fused the learned detail information and contextual features to perform semantic segmentation. Inspired by the above described methods, we design a novel lightweight network which integrates an edge guidance branch to produce edge attention features to guide the task of depth estimation.  \par
\section{Methodology}
This study proposes a novel lightweight monocular depth estimation network, EGD-Net. Fig. \ref{fig:pipeline} illustrates the architecture of EGD-Net, which is composed of four parts: multi-scale feature extractor, edge guidance branch, feature aggregation module and decoder. \par
\subsection{Multi-scale Feature Extractor}
The multi-scale feature extractor (MSFE) consists of a backbone (E-Block-1, 2, 3, 4) and an extension module. We adopt a lightweight CNN, MobileNetV2 \cite{sandler2018mobilenetv2} as the backbone. To adapt MobileNetV2 to the MDE task, we remove the layers after the fourth convolutional stage. Thus, the final features from the backbone are 1/16 size of the input image. To further enhance the representation ability of produced features, we add an extension module after the backbone. The extension module is comprised of six dilated Inverted Residual Blocks (IRBs) and each IRB has different dilation rate to capture multiple scales context information. Specifically, the dilation rates are set as 1, 2, 3, 1, 2 and 3. Furthermore, the expansion factor and stride values in IRB are set as 4 and 1 respectively. For convenience, we define the output feature maps from the MSFE as $D_{1}, D_{2}, D_{3}, D_{4}, D_{5}$, with strides of $2^1, 2^2, 2^3, 2^4, 2^4$, respectively. \par
\begin{figure}
	\centering
    \includegraphics[width=.45\textwidth]{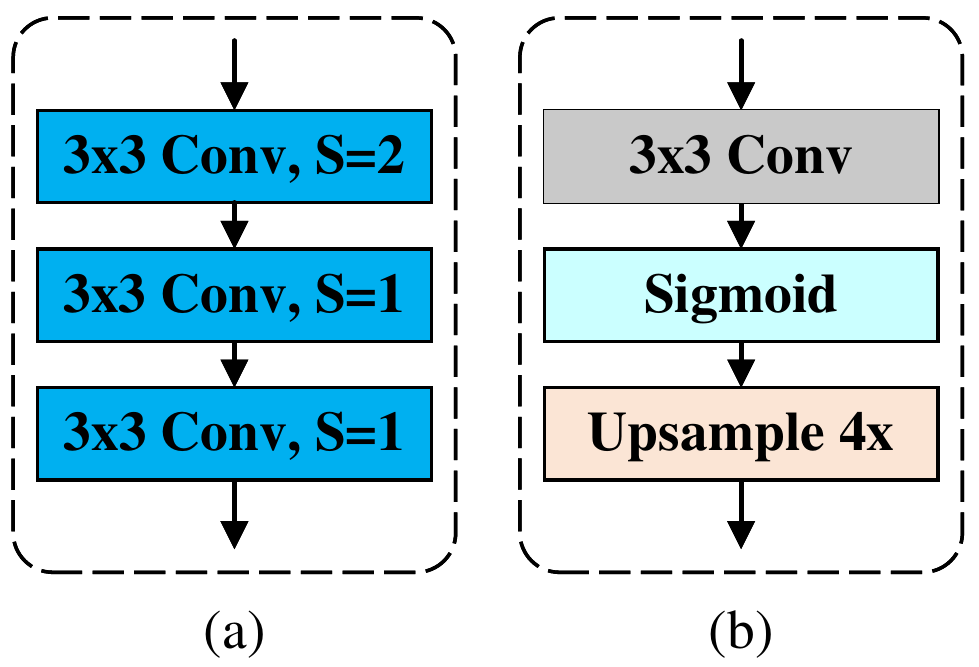}
    \caption{(a) The edge compact module, where S denotes stride. (b) The edge head, where ``$3\times3$ Conv" has no batch normalization and ReLU.}
    \label{fig:edge_input_output}
\end{figure}
\subsection{Edge Guidance Branch}
Edge information are important cues for CNNs to predict depth \cite{dijk2019neural, hu2019visualization}. To model the edge attention features for guiding the task of depth estimation, we design an edge guidance branch (EGB) which is composed of a few convolutional layers interleaved with channel attention-based feature fusion modules (CAFF). EGB takes image gradients as well as the intermediate features from the MSFE as input and outputs edge images and high resolution feature maps. We first extract image gradients in $x$ and $y$ directions from the input image with the Sobel operator. Image gradients are processed by an edge compact module (Fig. \ref{fig:edge_input_output} (a)) to generate features ($E_{1}$) at $1/2$ size of the input image. The generated features ($E_{1}$) are fused with $D_{1}$ through the CAFF. \par
\begin{figure}
	\centering
    \includegraphics[width=.45\textwidth]{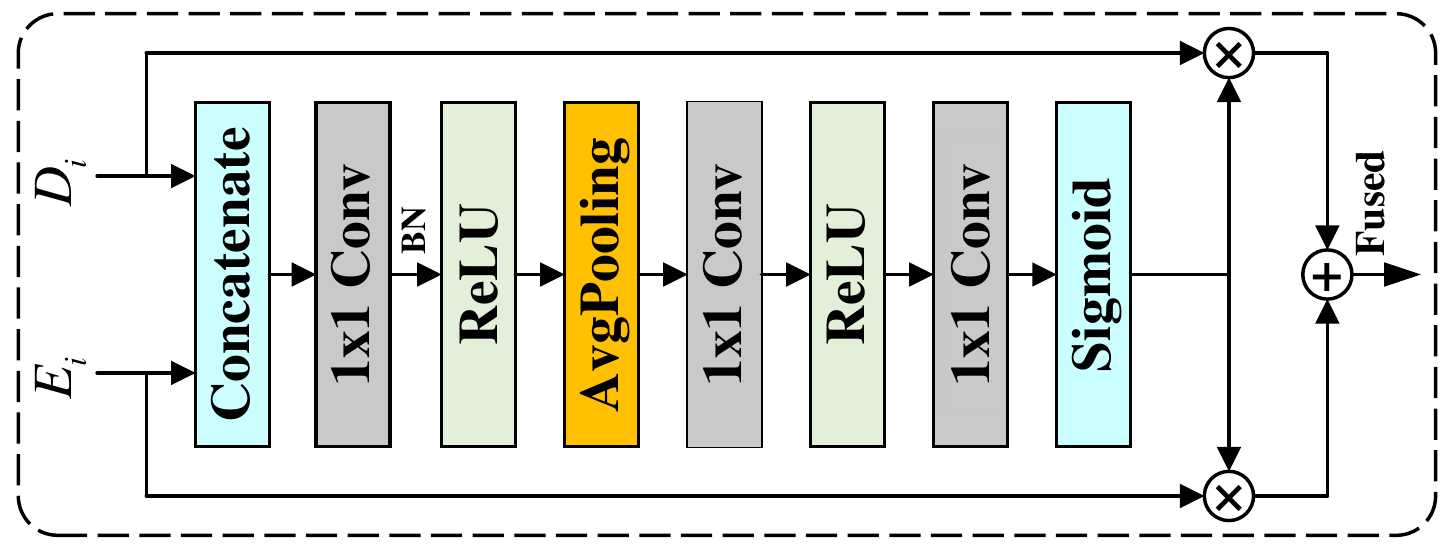}
    \caption{Illustration of the channel attention-based feature fusion module (CAFF), where ``$1\times1$ Conv" has no batch normalization and ReLU, ``$\times$" and ``$+$" represent multiply and add operations respectively.}
    \label{fig:feature_fusion}
\end{figure}

 
As shown in Fig. \ref{fig:feature_fusion}, intermediate feature maps generated by the MSFE ($D_{i}$) and EGB ($E_{i}$) are concatenated along the channel dimension. The concatenated features are passed through a $1\times1$ Conv+BN+ReLU layer to reduce the channel dimension to half size. Next, we pool the reduced features to a feature vector, which is then processed by $1\times1$ convolutions, ReLU and Sigmoid operators to get an attention map. The attention map is multiplied with $D_{i}$ and $E_{i}$ respectively. These multiplied features are added element-wise to build a fused feature. The fused feature is then convolved with a $3\times3$ Conv-BN-ReLU layer. We denote the feature maps generated by the EGB as $E_{1}, E_{2}, E_{3}, E_{4}, E_{5}, E_{6}$, with strides of $2^1, 2^1, 2^2, 2^3, 2^4, 2^4$, respectively. \par
Feature maps from EGB ($E_{3}, E_{4}, E_{5}$) are resized to 1/2 size of the input image and concatenated with $E_{2}$ and then pass through a $3\times3$ Conv-BN-ReLU layer. As shown in Fig. \ref{fig:pipeline}, the output feature ($F_{c}$) from the $3\times3$ Conv-BN-ReLU layer is used as the input of two directions. The first direction is the edge head, which produces edge maps for supervision. The second direction is passed to the decoder to act as the edge guidance feature. In addition, $E_{5}$ is convolved with a $3\times3$ Conv-BN-ReLU layer and generates $E_{6}$, which is fed to the TRFA module to aggregate with the context rich feature from the MSFE. In this study, the edge detection is modeled as a binary segmentation task. To obtain the ground-truth binary edge images, we adopt the Laplacian operator to extract edge maps from the ground-truth depth maps. The edge guidance branch is directly supervised by the binary edge labels. Thus, it learns edge attention features. \par
\begin{figure}
	\centering
    \includegraphics[width=.45\textwidth]{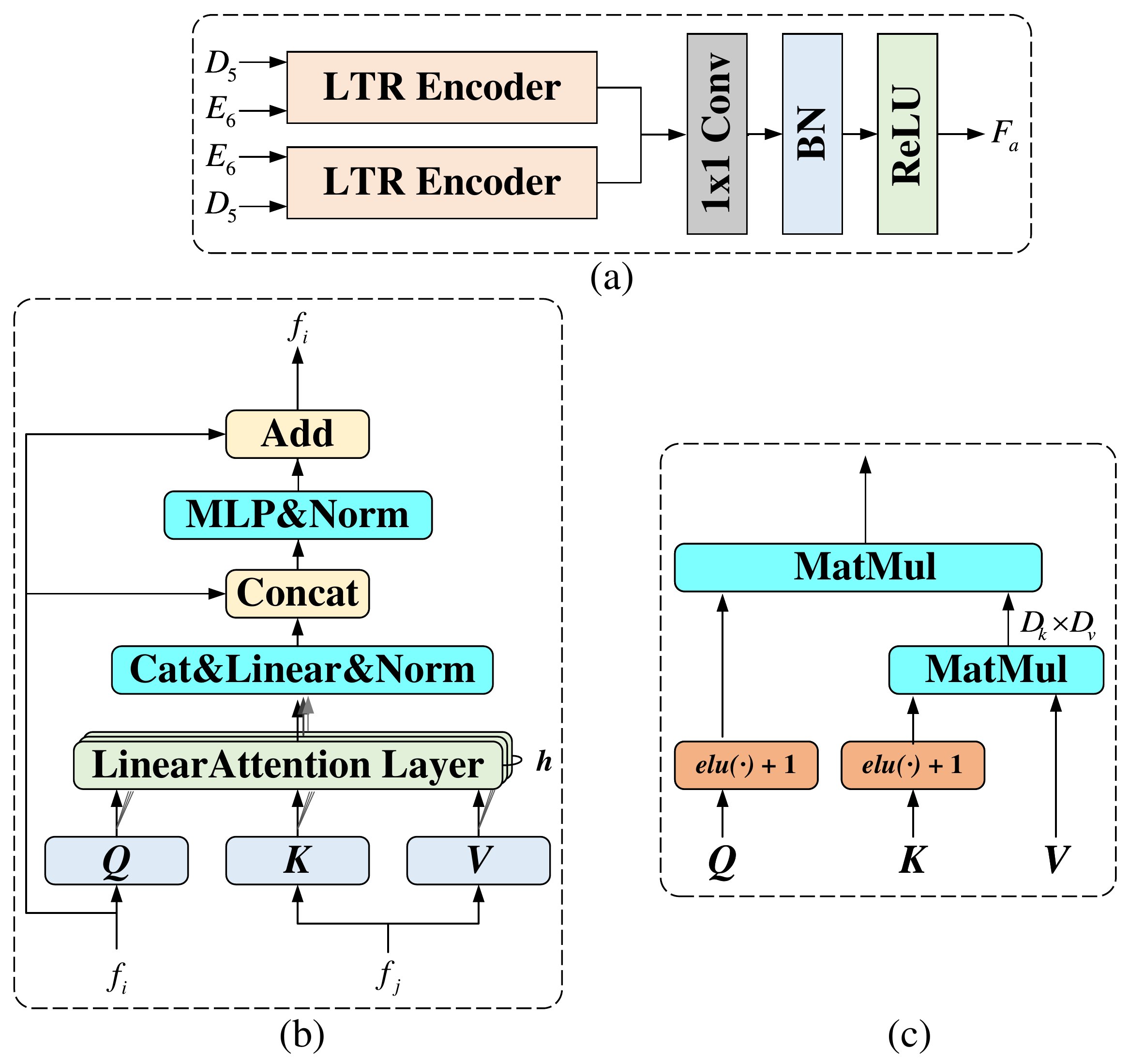}
    \caption{(a) The transformer-based feature aggregation module (TFAM), where LTR represents linear transformer. (b) LTR encoder layer, $h$ means the multiple heads of attention, which is set as 4 in this study. (c) Linear attention layer.}
    \label{fig:tfa_module}
\end{figure}
\subsection{Transformer-Based Feature Aggregation Module}
In order to combine the edge attention features from the EGB and context rich features from the MSFE to produce high resolution depth maps, we design a transformer-based feature aggregation module (TRFA). TRFA consists of two linear transformer encoder layers \cite{katharopoulos2020transformers} and a $1\times1$ Conv-BN-ReLU layer. The core element of the linear transformer encoder layer is that the linear attention layer computes the attention between a set of query vectors ($Q$) and key vectors ($K$) using dot-product similarity, which is then used to weigh a set of value vectors ($V$). Thus, the attention computation selects the relevant information through measuring the similarity between the query vector and each key vector. \par
Inspired by \cite{sun2021loftr}, we adopt the linear transformer encoder layer to capture the long-range dependencies (or global context) between the edge and context features through cross-attention in two directions. As shown in Fig. \ref{fig:tfa_module}, the input features to linear transformer encoders are $(D_{5}, E_{6})$ and $(E_{6}, D_{5})$ respectively. Then, features from the linear transformer encoder layers are concatenated and passed through the $1\times1$ Conv-BN-ReLU layer to aggregate them together. \par
The output features from TRFA are first upsampled to two times in size and then fed to the decoder. The decoder is composed of a $3\times3$ Conv-BN-ReLU layer and three decoder blocks (D-Block-1, 2, 3), each decoder block includes a $3\times3$ Conv-BN-ReLU layer and a bilinear interpolation with a scale factor of 2. The decoder incorporates features ($F_{a}$) from the TRFA, low-level features ($D_{1}, D_{2}, D_{3}$) from the MSFE, and the high resolution features ($F_{c}$) from the EGB to produce depth maps with the same size as input images. \par
\subsection{Loss Function}
\label{sect4:loss}
The designed network includes two branches that output depth maps and edge maps respectively. For the task of monocular depth estimation, we adopt the loss function proposed in \cite{dong2021mobilexnet}. This loss stacks the regular $L_{1}$ loss: ${{L}_{1}(d, d^{*}) = \frac{1}{N} \sum_{i}^{N} |d_{i} - d_{i}^{*}|}$ and the image gradient based $L_{1}$ loss:
${L_{grad}(d, d^{*}) = \frac{1}{N} \sum_{i}^{N} |{\bigtriangledown}_{x}(d_{i}, d_{i}^{*})| + |{\bigtriangledown}_{y}(d_{i}, d_{i}^{*})|}$, where $N$ is the total number of pixels being considered, $d$ and $d^{*}$ are the predicted and ground-truth depth, ${\bigtriangledown}_{x}$ and ${\bigtriangledown}_{y}$ are the spatial derivatives in $x$ and $y$ directions. The depth estimation loss can be written as: $L_{depth} = L_{1} + L_{grad}$. For edge detection, we employ the standard binary cross-entropy (BCE) loss $L_{edge}$, which is defined as:$L_{edge} = -{\sum_{i}(e^{*}_{i}log{e_{i} + (1-e^{*}_{i})log(1-e_{i})})}$, where $e_{i}$ and $e^{*}_{i}$ are the detected and ground-truth edges. Finally, the whole loss function is formulated as: $L = {\lambda}_{1}{L_{depth}} + {\lambda}_{2}{L_{edge}}$,
where ${\lambda}_{1}$ and ${\lambda}_{2}$ are the hyper-parameters, we empirically set ${\lambda}_{1}$ = 1 and ${\lambda}_{2}$ = 20.

\section{Experiments}
To demonstrate the effectiveness of our proposed EGD-Net, we evaluate it on the NYU depth v2 dataset \cite{silberman2012indoor}.

\subsection{Implementation Details}
\label{sect4:implement}
The designed network is implemented in PyTorch. A workstation with a single Nvidia RTX 3090 GPU is used for training and testing. The weights of the backbone of the MSFE are initialized with the weights pre-trained on ImageNet. The other layers are randomly initialized. The training is optimized by using the SGD optimizer, and the batch size is set as 8. We train the network for 25 epochs. The poly learning rate policy is adopted, the learning rate for the $n^{th}$ epoch is $init\underline{~}lr \times (1 - \frac{n}{max\underline{~}epoch})^{power}$, where the $init\underline{~}lr$ and power are set as 0.01 and 0.9 respectively. \par 
During training, we employ data augmentation approaches to increase the diversity of training samples. Data augmentations are applied to each RGB and ground-truth depth image pair in an online fashion:
\begin{itemize}
    \item {Random Flips:} RGB and ground-truth depth image pairs are horizontally flipped at a probability of 0.5.
    \item {Random Rotation:} RGB and ground-truth depth image pairs are randomly rotated by a degree of $r \in [-5, 5]$.
    \item {Color Jitter:} the brightness, contrast and saturation values of the RGB images are randomly scaled by a factor of $c \in [0.6, 1.4]$.
\end{itemize}
\begin{table*}[!ht]
	\caption{Comparison of performances on the NYU depth v2 dataset \cite{silberman2012indoor}. $\uparrow$ means higher is better, $\downarrow$ means lower is better. The \textcolor{red}{red} and \textcolor{red}{\textbf{bold}} values indicate the best results.}
	\centering
	\label{table: comparison_on_nyu}
	\begin{tabular}{ c | c | c | c | c | c | c | c } 
		\hline
		Method & Backbone & Params. $\downarrow$ & RMSE $\downarrow$ & REL $\downarrow$ & {${{\delta}_1} \uparrow $} & {${{\delta}_2} \uparrow $} & {${{\delta}_3} \uparrow $}  \\
		\hline\hline
%		Laina et al. \cite{laina2016deeper} & ResNet-50 & 63.4 M & 0.573 & 0.127 & 0.811 & 0.953 & 0.988 \\
		Hu et al. \cite{hu2019revisiting} & SENet-154 & 157.0 M & 0.530 & 0.115 & 0.866 & 0.975 & 0.993 \\
		Chen et al. \cite{chen2019structure} & SENet-154 & 210.3 M & 0.514 & \textbf{\textcolor{red}{0.111}} & \textbf{\textcolor{red}{0.878}} & \textbf{\textcolor{red}{0.977}} & \textbf{\textcolor{red}{0.994}} \\
		Wofk et al. \cite{wofk2019fastdepth} & MobileNet & 20.67 M & 0.529 & 0.155 & 0.789 & 0.950 & 0.987 \\
		Tu et al. \cite{tu2020efficient} & MobileNetV2 & 5.7 M & 0.531 & 0.147 & 0.801 & 0.956 & 0.989 \\ 
		Rudolph et al. \cite{rudolph2022lightweight} & DDRNet-23-slim & 5.8 M & 0.501 & 0.138 & 0.823 & 0.961 & 0.990 \\
		MobileXNet \cite{dong2021mobilexnet} & MobileNet & 24.95 M & 0.507 & 0.149 & 0.807 & 0.953 & 0.989 \\ 
		Ours & MobileNetV2 & \textbf{\textcolor{red}{2.21 M}} & \textbf{\textcolor{red}{0.486}} & 0.136 & 0.825 & 0.960 & 0.990  \\ \hline
	\end{tabular}
\end{table*}
\begin{table*}[!ht]
	\caption{Ablation study on different feature aggregation methods. $\uparrow$ means higher is better, $\downarrow$ means lower is better. ``w CAFFM" means the EGB includes the CAFF module. ``w/o CAFFM" indicates the EGB does not include the CAFF module. The \textcolor{red}{red} and \textcolor{red}{\textbf{bold}} values indicate the best results.}
	\centering
	\label{table: ablation_component}
	\begin{tabular}{ c | c | c | c | c | c | c } 
		\hline
		Method & Params. $\downarrow$ & RMSE $\downarrow$ & REL $\downarrow$ & {${{\delta}_1} \uparrow $} & {${{\delta}_2} \uparrow $} & {${{\delta}_3} \uparrow $} \\
		\hline\hline
		Baseline & \textbf{\textcolor{red}{1.56 M}} & 0.520 & 0.149 & 0.796 & 0.955 & 0.988 \\ 
		Baseline + EGB & 2.01 M & 0.514 & 0.149 & 0.808 & 0.959 & \textbf{\textcolor{red}{0.990}} \\
		Baseline + TAFM & 2.19 M & 0.506 & 0.144 & 0.810 & 0.955 & 0.986 \\
		Baseline + EGB (w CAFF) + TAFM & 2.21 M & \textbf{\textcolor{red}{0.486}} & \textbf{\textcolor{red}{0.136}} & \textbf{\textcolor{red}{0.825}} & \textbf{\textcolor{red}{0.960}} & \textbf{\textcolor{red}{0.990}} \\ 
		Baseline + EGB (w/o CAFF) + TAFM & 2.16 M & 0.489 & 0.144 & 0.817 & \textbf{\textcolor{red}{0.960}} & 0.989 \\ \hline 
	\end{tabular}
\end{table*}
\begin{table*}[h]
	\caption{Ablation study on different backbones. $\uparrow$ means higher is better, $\downarrow$ means lower is better. The \textcolor{red}{red} and \textcolor{red}{\textbf{bold}} values indicate the best results.}
	\centering
	\label{table: comparison_backbones}
	\begin{tabular}{ c | c | c | c | c | c | c } 
		\hline
		Backbone & Params. $\downarrow$ & RMSE $\downarrow$ & REL $\downarrow$ & {${{\delta}_1} \uparrow $} & {${{\delta}_2} \uparrow $} & {${{\delta}_3} \uparrow $} \\
		\hline\hline
		ResNet-18 \cite{he2016deep} & 6.43 M & 0.489 & 0.144 & 0.818 & \textbf{\textcolor{red}{0.960}} & \textbf{\textcolor{red}{0.990}} \\
		EffcientNet-B0 \cite{tan2019efficientnet} & 2.63 M & 0.555 & 0.161 & 0.771 & 0947 & 0.988 \\		
		ShuffleNetV2 \cite{ma2018shufflenet} & 3.37 M & 0.514 & 0.155 & 0.798 & 0.949 & 0.986 \\
		MobileNetV2 \cite{sandler2018mobilenetv2} & \textbf{\textcolor{red}{2.21 M}} & \textbf{\textcolor{red}{0.486}} & \textbf{\textcolor{red}{0.136}} & \textbf{\textcolor{red}{0.825}} & \textbf{\textcolor{red}{0.960}} & \textbf{\textcolor{red}{0.990}} \\ \hline
	\end{tabular}
\end{table*}
\subsection{Dataset and Evaluation Metric}
We evaluate the proposed method on the commonly used NYU depth v2 dataset \cite{silberman2012indoor}, which was collected in real-world indoor surroundings with a Microsoft Kinect camera. The original images have a resolution of $640\times480$ pixels. In this work, we train our method on the training set proposed by Hu et al. \cite{hu2019revisiting} and evaluate it on the offical testing set including 654 RGB and depth image pairs. Each image pair is downsampled to $342\times256$ and then center cropped to $320\times240$. We first compare our proposed method with state-of-the-art methods and then perform ablation experiments to validate the contribution of each component of the proposed network. \par
We adopt three widely used metrics to evaluate the proposed method. Since our aim is to estimate depth maps from RGB images, only the error and accuracy metrics of depth estimation are compared. Let $N$ denotes the total number of valid pixels, $d_{i}$ and $d_{i}^{*}$ represent the estimated and ground-truth depth values at the pixel indexed by $i$, respectively. The error metrics are defined as follows:
\begin{itemize}
    \item {Root Mean Square Error (RMSE):} $\sqrt{\frac{1}{N}\sum_{i}^{N}|d_{i} - d_{i}^{*}|^2}$.
    \item {Mean Relative Error (REL):} ${ \frac{1}{N}\sum_{i}^{N}\frac{|d_{i} - {d_{i}^{*}}|}{{d_{i}}}}$.
    \item {$\delta_{i}$ Accuracy:} \% of $d_{i}$ s.t. ${max(\frac{d^{*}}{d}, \frac{d}{d^{*}}) < \delta_{i}, \delta_{i} = 1.25^{i}}$.
    %
%    \item {Running time ($t_{GPU}$):} the average execution time of testing each frame on a single less powerful GPU, the GTX 1080 GPU.
\end{itemize}
\subsection{Comparison with State-of-the-art}
In this subsection, we compare the performance of our EGD-Net with state-of-the-art methods \cite{hu2019revisiting, chen2019structure, wofk2019fastdepth, tu2020efficient, rudolph2022lightweight, dong2021mobilexnet} in terms of the amount of network parameters (Params., million), error (RMSE and REL) and accuracy ($\delta_{1}$, $\delta_{2}$ and $\delta_{3}$) metrics. Quantitative results of our proposed method and state-of-the-art methods are listed in Table \ref{table: comparison_on_nyu}. For \cite{hu2019revisiting, chen2019structure} we report the corresponding results from their papers. The results of \cite{tu2020efficient, rudolph2022lightweight} are reported in \cite{rudolph2022lightweight}. We retrained \cite{wofk2019fastdepth, dong2021mobilexnet} with the same training and testing procedure as described in Section \ref{sect4:implement}. To compare fairly, \cite{wofk2019fastdepth, dong2021mobilexnet} are supervised by the depth loss described in Section \ref{sect4:loss}. \par
It can be observed that: (1) among all methods, EGD-Net has the lowest amount of parameters. In particular, EGD-Net has $>$2.5$\times$ fewer parameters than \cite{tu2020efficient, rudolph2022lightweight}, $>$11$\times$ fewer parameters than MobileXNet \cite{dong2021mobilexnet}, $>$71$\times$ fewer parameters than Hu et al. \cite{hu2019revisiting}, and $>$95$\times$ fewer than Chen et al. \cite{chen2019structure}; (2) with much fewer network parameters, EGD-Net generates the best RMSE performance while its $\delta_{2}$ and $\delta_{3}$ metrics are very close to Hu et al. \cite{hu2019revisiting} and Chen et al. \cite{chen2019structure}; (3) EGD-Net outperforms \cite{wofk2019fastdepth, tu2020efficient, rudolph2022lightweight, dong2021mobilexnet} in terms of all error and accuracy metrics. \par
\begin{figure*}
	\centering
    \includegraphics[width=.85\textwidth]{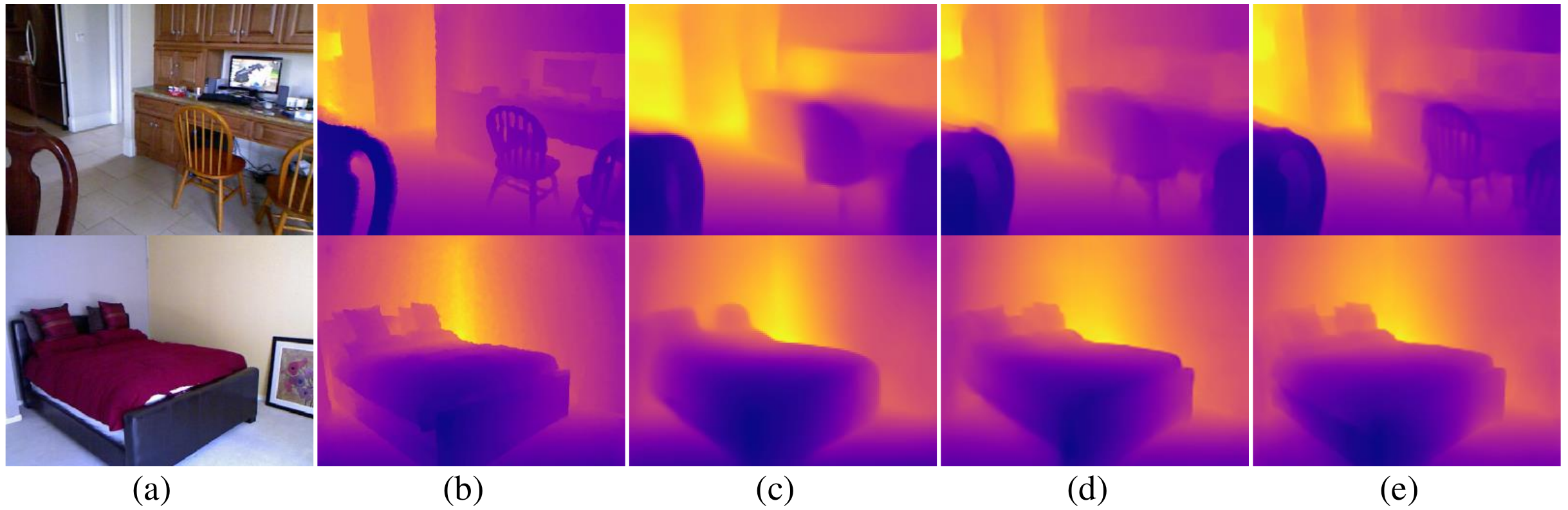}
    \caption{Qualitative results from the NYU depth v2 dataset. (a) RGB image, (b) Ground-truth depth, (c) Wofk et al. \cite{wofk2019fastdepth}, (d) MobileXNet \cite{dong2021mobilexnet} and (e) Our results. Color represents depth (yellow is far, blue is close).}
    \label{fig:qualitative_nyu}
\end{figure*}

Regarding the running speed, EGD-Net runs about 96 fps (GPU reference time is 10.4 ms) on a Nvidia GTX 1080 GPU (2580 CUDA cores and 8GB memory), which is adequate for real-time robotic applications. We present the qualitative comparison of our results with Wofk et al. \cite{wofk2019fastdepth} and MobileXNet \cite{dong2021mobilexnet} in Fig. \ref{fig:qualitative_nyu}. It can be observed that our proposed method can predict finely detailed object boundaries while \cite{wofk2019fastdepth, dong2021mobilexnet} cannot predict clearly. \par 
\subsection{Ablation Experiments and Analyses}
To analyze the contribution of each component of the designed network, we perform experiments with different deployments on the NYU Depth v2 dataset \cite{silberman2012indoor}. The training and testing strategies are kept the same as Section \ref{sect4:implement}. \par
\subsubsection{Contribution of Different Components}
We setup a baseline network that is consisted of the multi-scale feature extractor and the decoder shown in Fig. \ref{fig:pipeline}. The baseline method is trained with the depth loss described in Section \ref{sect4:loss}. The edge guidance branch and transformer-based feature aggregation module are added to the baseline step by step. \par
As show in Table \ref{table: ablation_component}, the baseline has the lowest amount of parameters, while it yields the worst results. When we combine the proposed EGB with the baseline, it outperforms the baseline in terms of RMSE, $\delta_{1}$, $\delta_{2}$ and $\delta_{3}$. To explore the influence of the proposed TRFA, we append it to the baseline network and train it with the depth loss. With the same training procedure, it outperforms the baseline and the combination of ``Baseline + EGB". Finally, our whole EGD-Net (line 4, Table \ref{table: ablation_component}), with both EGB and TRFA and trained with the whole loss function, yields the best performance in terms of all metrics. To evaluate the contribution of the proposed CAFF, we design a variant by replacing CAFF with pixel-wise addition (Baseline + EGB (w/o CAFF) + TAFM). According to the last two lines, the CAFF improves the performance of EGD-Net in term of all error and accuracy metrics. \par
\subsubsection{Comparison of Different Backbones}
In this subsection, we investigate the influence of adopting different backbones in the MSFE. We compare MobileNetV2 \cite{sandler2018mobilenetv2} with three CNNs, ResNet-18 \cite{he2016deep}, EfficientNet-B0 \cite{tan2019efficientnet} and ShuffleNetV2 \cite{ma2018shufflenet}. Specifically, ResNet-18 \cite{he2016deep} is a general CNN, ShuffleNetV2 \cite{ma2018shufflenet} and EfficientNet-B0 \cite{tan2019efficientnet} are lightweight CNNs. The weights of all backbones are initialized from the pre-trained models on ImageNet. To make the backbones compatible with the fixed feature aggregation module, we fixed the channel dimension of the final feature maps from the MSFE ($D_{5}$) and EGB ($E_{6}$) to 128. We report the results of four backbones in Table \ref{table: comparison_backbones}. As can be observed, when using MobileNetV2 \cite{sandler2018mobilenetv2} as backbone the proposed method achieves the best trade-off between accuracy and computation complexity. In particular, it has the lowest amount of parameters and yields the best performance in terms of both error and accuracy metrics. \par
\section{Conclusion}
In this paper, we introduced a novel lightweight monocular depth estimation network, named EGD-Net. Specifically, we designed an Edge Guidance Branch to detect edges and produce edge attention features that contain edge information. Moreover, a transformer-based feature aggregation module has been designed to learn the long-range dependencies between the edge and context features and aggregate them together. Extensive experiments on the NYU depth v2 dataset demonstrated the effectiveness of our proposed network. In future work, we will extend our approach to real-world and simulated outdoor environments.
\bibliographystyle{IEEEtran}
\bibliography{references.bib}
\end{document}